%% file: paper.tex
\documentclass[10pt,twocolumn,letterpaper]{article}

\usepackage{cvpr}
\usepackage{times}
\usepackage{epsfig}
\usepackage{graphicx}
\usepackage{amsmath}
\usepackage{amssymb}

\usepackage{color}
\usepackage{url}            % simple URL typesetting
\usepackage{amsfonts}       % blackboard math symbols
\usepackage{microtype}      % microtypography
\usepackage{floatrow}

\newfloatcommand{capbtabbox}{table}[][\FBwidth]

\newcommand{\multiline}[1]{{\begin{tabular}{@{}c@{}} #1 \end{tabular}}}

\usepackage[pagebackref=true,breaklinks=true,letterpaper=true,colorlinks,bookmarks=false]{hyperref}

\cvprfinalcopy % *** Uncomment this line for the final submission
\def\cvprPaperID{120} % *** Enter the CVPR Paper ID here

\begin{document}

\title{A Hierarchical Approach for Generating Descriptive Image Paragraphs}

\author{
  Jonathan Krause \quad Justin Johnson \quad Ranjay Krishna \quad Li Fei-Fei \\
  Stanford University \\
  {\tt\small \{jkrause,jcjohns,ranjaykrishna,feifeili\}@cs.stanford.edu}
}

\maketitle

\begin{abstract}
\input{abstract.tex}
\end{abstract}

\section{Introduction}
\label{sec:intro}
\input{intro.tex}

\section{Related Work}
\label{sec:related}
\input{related.tex}

\section{Paragraphs are Different}
\label{sec:paragraphs}
\input{paragraphs.tex}

\section{Method}
\label{sec:method}
\input{method.tex}

\section{Experiments}
\label{sec:experiments}
\input{experiments.tex}

\section{Conclusion}
\label{sec:discussion}
\input{discussion.tex}

{\small
\bibliographystyle{ieee}
\bibliography{references}
}

\end{document}

%% file: abstract.tex
Recent progress on image captioning has made it possible to generate novel
sentences describing images in natural language, but compressing an image into
a single sentence can describe visual content in only coarse detail.  While
one new captioning approach, dense captioning, can potentially describe images
in finer levels of detail by captioning many regions within an image, it in
turn is unable to produce a coherent story for an image.  In this paper we
overcome these limitations by generating entire paragraphs for describing
images, which can tell detailed, unified stories.  We develop a model that
decomposes both images and paragraphs into their constituent parts, detecting
semantic regions in images and using a hierarchical recurrent neural network to
reason about language.  Linguistic analysis confirms the complexity
of the paragraph generation task, and thorough experiments on a new dataset
of image and paragraph pairs demonstrate the effectiveness of our approach.

%% file: intro.tex
Vision is the primary sensory modality for human perception, and language is
our most powerful tool for communicating with the world. Building systems that
can simultaneously understand visual stimuli and describe them in natural
language is therefore a core problem in both computer vision and artificial
intelligence as a whole.  With the advent of large datasets pairing images with
natural language descriptions~\cite{lin2014microsoft,young2014image,hodosh2013framing,krishnavisualgenome} it
has recently become possible to generate novel sentences describing
images~\cite{chen2015mind,donahue2015long,karpathy2015deep,mao2014deep,vinyals2015show}.
While the success of these methods is encouraging, they all share one key
limitation: \emph{detail}.  By only describing images with a single high-level
sentence, there is a fundamental upper-bound on the quantity and quality of
information approaches can produce.

\begin{figure}
\includegraphics[width=0.95\linewidth]{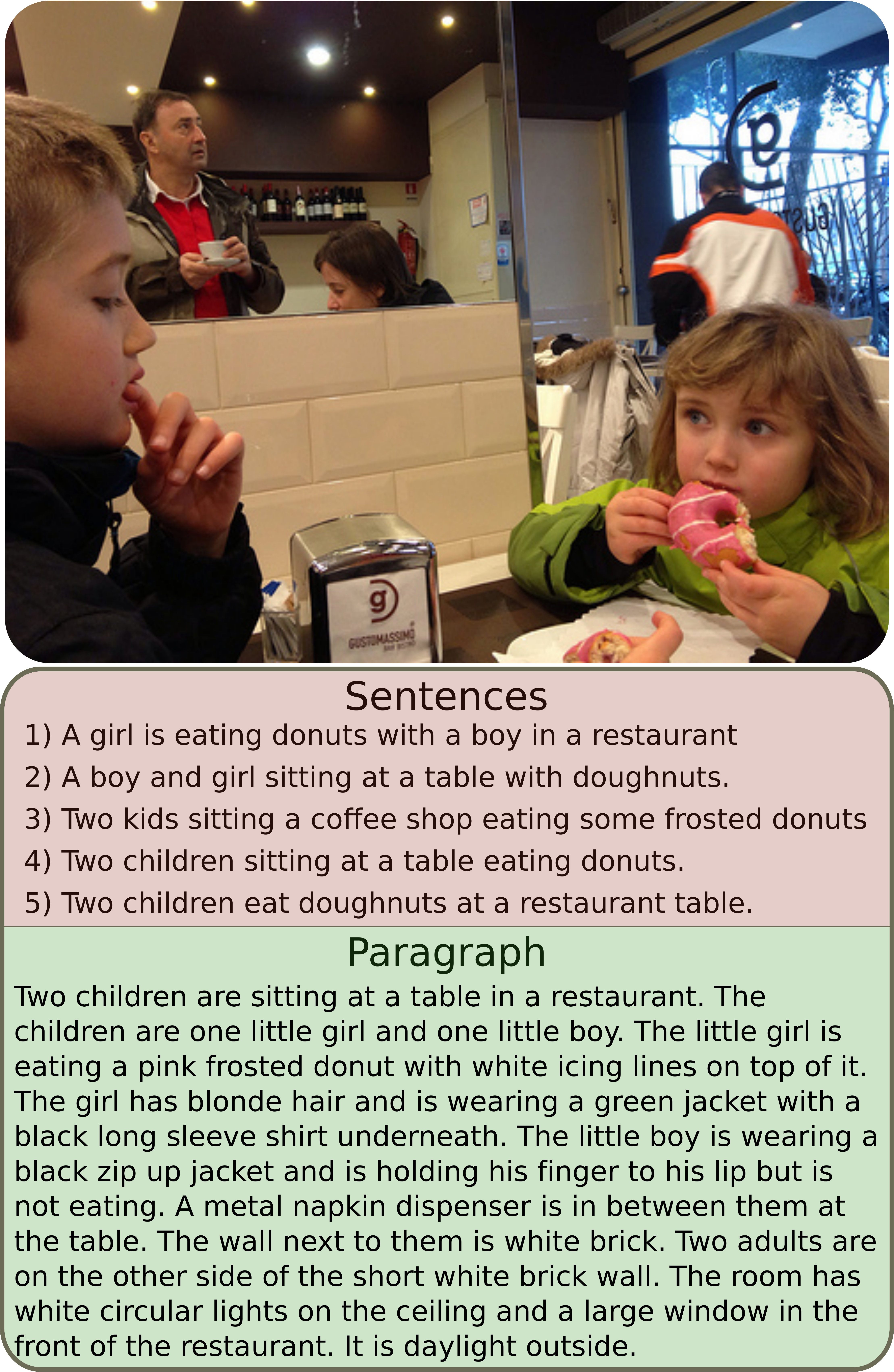}
{\caption{
Paragraphs are longer, more informative, and more linguistically complex than sentence-level captions.
Here we show an image with its sentence-level captions from MS COCO~\cite{lin2014microsoft} (top) and the paragraph used in this work (bottom).
\vspace{-3mm}
}
\label{fig:coco_paragraph}
}
\end{figure}

One recent alternative to sentence-level captioning is the task of dense
captioning~\cite{johnson2016densecap}, which overcomes this limitation by
detecting many regions of interest in an image and describing each with a short
phrase.  By extending the task of object detection to include natural language
description, dense captioning describes images in considerably more detail than
standard image captioning.  However, this comes at a cost:
descriptions generated for dense captioning are not coherent, \ie they do not
form a cohesive whole describing the entire image.

In this paper we address the shortcomings of both traditional image captioning
and the recently-proposed dense image captioning by introducing the task of
generating paragraphs that richly describe images
(Fig.~\ref{fig:coco_paragraph}).  Paragraph generation combines the strengths
of these tasks but does not suffer from their weaknesses -- like traditional
captioning, paragraphs give a coherent natural language description for images,
but like dense captioning, they can do so in fine-grained detail.

Generating paragraphs for images is challenging, requiring both fine-grained
image understanding and long-term language reasoning. To overcome these
challenges, we propose a model that decomposes images and paragraphs into their
constituent parts: We break images into semantically meaningful pieces by
detecting objects and other regions of interest, and we reason about language
with a hierarchical recurrent neural network, decomposing paragraphs into their
corresponding sentences.
In addition, we also demonstrate for the first time the ability to transfer
visual and linguistic knowledge from large-scale region
captioning~\cite{krishnavisualgenome}, which we show has the ability to improve
paragraph generation.

To validate our method, we collected a dataset of image and paragraph pairs,
which complements the whole-image and region-level annotations of
MS COCO~\cite{lin2014microsoft} and Visual Genome~\cite{krishnavisualgenome}.
To validate the complexity of the paragraph generation task, we performed a
linguistic analysis of our collected paragraphs, comparing them to
sentence-level image captioning.  We compare our approach with numerous
baselines, showcasing the benefits of hierarchical modeling for generating
descriptive paragraphs.

The rest of this paper is organized as follows: Sec.~\ref{sec:related}
overviews related work in image captioning and hierarchical RNNs,
Sec.~\ref{sec:paragraphs} introduces the paragraph generation task, describes
our newly-collected dataset, and performs a simple linguistic analysis on it,
Sec.~\ref{sec:method} details our model for paragraph generation,
Sec.~\ref{sec:experiments} contains experiments, and Sec.~\ref{sec:discussion}
concludes with discussion.

%% file: related.tex
\paragraph{Image Captioning}
Building connections between visual and textual data has been a longstanding
goal in computer vision. One line of work treats the problem as a ranking task,
using images to retrieve relevant captions from a database and
vice-versa~\cite{farhadi2010every,hodosh2013framing,karpathy2014deep}. Due to
the compositional nature of language, it is unlikely that any database will
contain all possible image captions; therefore another line of work
focuses on generating captions directly.
Early work uses handwritten
templates to generate language~\cite{kulkarni2011baby} while more recent
methods train recurrent neural network language models conditioned on image
features~\cite{chen2015mind,donahue2015long,karpathy2015deep,mao2014deep,vinyals2015show,you2016image}
and sample from them to generate text.  Similar methods have also been applied
to generate captions for
videos~\cite{donahue2015long,yao2015describing,yu2016video}.

A handful of approaches to image captioning reason not only about whole images
but also image regions. Xu \etal~\cite{xu2015show} generate captions using a
recurrent network with attention, so that the model
produces a distribution over image regions for each word.  In contrast to their work, which
uses a coarse grid as image regions, we use semantically meaningful
regions of interest.  Karpathy and Fei-Fei~\cite{karpathy2015deep} use a
ranking loss to align image regions with sentence fragments but do not do
generation with the model.  Johnson \etal~\cite{johnson2016densecap} introdue
the task of dense captioning, which detects and describes regions of interest,
but these descriptions are independent and do not form a coherent whole.

There has also been some pioneering work on video captioning with multiple sentences~\cite{rohrbach2014coherent}.
While videos are a natural candidate for multi-sentence description generation, image captioning cannot leverage strong temporal dependencies, adding extra challenge.

\vspace{-10mm}

\paragraph{Hierarchical Recurrent Networks}
In order to generate a paragraph description, a model must reason about
long-term linguistic structures spanning multiple sentences. Due to vanishing
gradients, recurrent neural networks trained with stochastic gradient descent
often struggle to learn long-term dependencies.
Alternative recurrent architectures such as long-short term memory
(LSTM)~\cite{hochreiter1997long} help alleviate this problem through a gating
mechanism that improves gradient flow. Another solution is a \emph{hierarchical}
recurrent network, where the architecture is designed such that different parts
of the model operate on different time scales.

Early work applied hierarchical recurrent networks to simple
algorithmic problems~\cite{el1995hierarchical}. The Clockwork
RNN~\cite{koutnik2014clockwork} uses a related technique for audio signal
generation, spoken word classification, and handwriting recognition; a similar
hierarchical architecture was also used in~\cite{chan2016listen} for speech
recognition. In these approaches, each recurrent unit is updated on a fixed
schedule: some units are updated on every timestep, while other units
might be updated every other or every fourth timestep. This type of
hierarchy helps reduce the vanishing gradient problem, but the hierarchy of the
model does not directly reflect the hierarchy of the output sequence.

More related to our work are hierarchical architectures that directly mirror
the hierarchy of language. Li \etal~\cite{li2015hierarchical} introduce a
hierarchical autoencoder, and Lin \etal~\cite{lin2015hierarchical} use
different recurrent units to model sentences and words.  Most similar to our
work is Yu \etal~\cite{yu2016video}, who generate multi-sentence descriptions
for cooking videos using a different hierarchical model.
Due to the less constrained non-temporal setting in our
work, our method has to learn in a much more generic fashion and has been made
simpler as a result, relying more on learning the interplay between sentences.
Additionally, our method reasons about semantic regions in images, which both
enables the transfer of information from these regions and leads to more
interpretability in generation.

%% file: paragraphs.tex
\begin{table}[t]
\capbtabbox{%
\begin{tabular}{ccc}
  %& \begin{tabular}{@{}c@{}}Visual \\ Genome\end{tabular} & MS COCO & Paragraphs \\
  \hline
  & \multiline{Sentences\\COCO~\cite{lin2014microsoft}} & \multiline{Paragraphs\\Ours}\\
  \hline
  Description Length & 11.30 & 67.50 \\
  Sentence Length & 11.30 & 11.91 \\
  Diversity & 19.01 & 70.49\\
  Nouns & 33.45\% & 25.81\%\\
  Adjectives & 27.23\% & 27.64\%\\
  Verbs & 10.72\% & 15.21\%\\
  Pronouns & 1.23\% & 2.45\%\\
  \hline
\end{tabular}
}{
  \caption{
Statistics of paragraph descriptions, compared with sentence-level captions used in prior work.
Description and sentence lengths are represented by the number of tokens present, diversity is the inverse of the average CIDEr score between sentences of the same image, and part of speech distributions are aggregated from Penn Treebank~\cite{marcus1993building} part of speech tags.
}
  \label{tab:data_stats}
}
\end{table}

To what extent does describing images with paragraphs differ from
sentence-level captioning?  To answer this question, we collected a novel
dataset of paragraph annotations, comparised of 19,551 MS
COCO~\cite{lin2014microsoft} and Visual Genome~\cite{krishnavisualgenome}
images, where each image has been annotated with a paragraph description.
Annotations were collected on Amazon Mechanical Turk, using U.S. workers with
at least 5,000 accepted HITs and an acceptance rate of 98\% or
greater\footnote{Available at \url{http://cs.stanford.edu/people/ranjaykrishna/im2p/index.html}}, and were additionally subject to
automatic and manual spot checks on quality.
Fig.~\ref{fig:coco_paragraph} demonstrates an example,
comparing our collected paragraph with the five corresponding sentence-level captions from MS COCO.
Though it is clear that the paragraph is longer and more descriptive than any
one sentence, we note further that a single paragraph can be more detailed than
\emph{all five} sentence captions, even when combined.  This occurs because of
redundancy in sentence-level captions -- while each caption might use slightly
different words to describe the image, since all sentence captions have the
goal of describing the image as a whole, they are fundamentally limited in
terms of both diversity and their total information.

We quantify these observations along with various other statistics of language
in Tab.~\ref{tab:data_stats}.  For example, we find that each paragraph is
roughly six times as long as the average sentence caption, and the individual
sentences in each paragraph are of comparable length as
sentence-level captions.  To examine the issue of sentence diversity, we
compute the average CIDEr~\cite{vedantam2015cider} similarity between COCO
sentences for each image and between the individual sentences in each collected
paragraph, defining the final diversity score as 100 minus the average CIDEr
similarity.  Viewed through this metric, the difference in diversity is
striking -- sentences within paragraphs are substantially more diverse than
sentence captions, with a diversity score of 70.49 compared to only 19.01.
This quantifiable evidence demonstrates that sentences in paragraphs provide
significantly more information about images.

Diving deeper, we performed a simple linguistic analysis on
COCO sentences and our collected paragraphs, comprised of annotating each word
with a part of speech tag from Penn Treebank via Stanford
CoreNLP~\cite{manning2014stanford} and aggregating parts of speech into
higher-level linguistic categories.  A few common parts of speech are given in
Tab.~\ref{tab:data_stats}.  As a proportion, paragraphs have somewhat more
verbs and pronouns, a comparable frequency of adjectives, and somewhat fewer
nouns.  Given the nature of paragraphs, this makes sense -- longer descriptions
go beyond the presence of a few salient objects and include information about
their properties and relationships.  We also note but do not quantify that
paragraphs exhibit higher frequencies of more complex linguistic phenomena, \eg
coreference occurring in Fig.~\ref{fig:coco_paragraph}, wherein sentences
refer to either ``two children'', ``one little girl and one little boy'', ``the
girl'', or ``the boy.'' We belive that these types of long-range phenomena are
a fundamental property of descriptive paragraphs with human-like language and
cannot be adequately explored with sentence-level captions.

%% file: method.tex
\begin{figure*}
  \centering
  \includegraphics[width=1.0\textwidth]{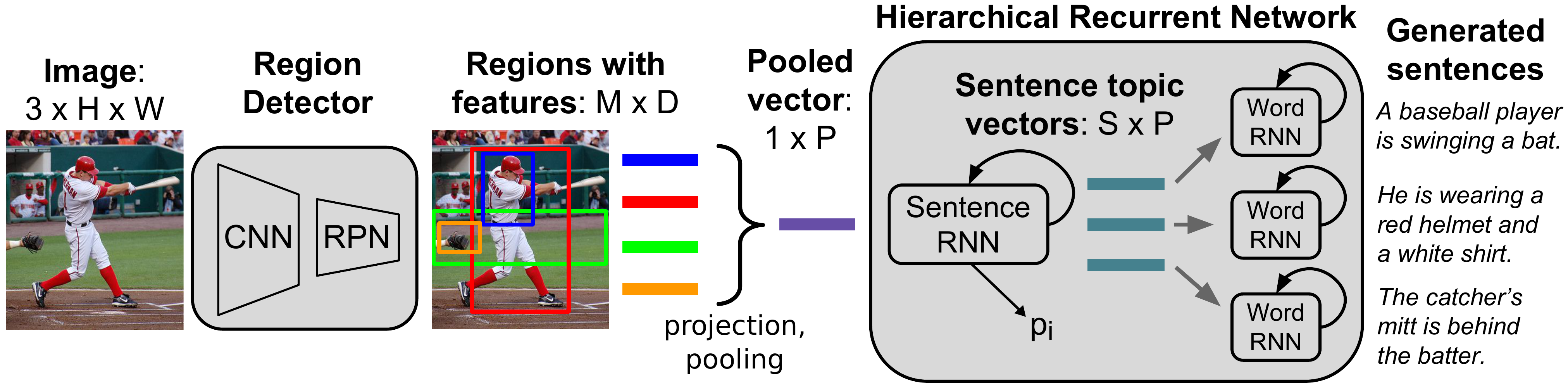}
  \caption{Overview of our model.
    Given an image (left), a region detector (comprising a convolutional
    network and a region proposal network) detects regions of interest and
    produces features for each. Region features are projected to $\mathbb{R}^P$,
    pooled to give a compact image representation, and passed to a hierarchical
    recurrent neural network language model comprising a sentence RNN and a
    word RNN.  The sentence RNN determines the number of sentences to generate
    based on the halting distribution $p_i$ and also generates sentence topic
    vectors, which are consumed by each word RNN to generate sentences.
		\vspace{-4mm}
  }
	\vspace{-2mm}
  \label{fig:system}
\end{figure*}

\paragraph{Overview}
Our model takes an image as input, generating a natural-language paragraph
describing it, and is designed to take advantage of the compositional structure
of both images and paragraphs.  Fig.~\ref{fig:system} provides an
overview.  We first decompose the input image by detecting objects and other
regions of interest, then aggregate features across these regions to produce a
pooled representation richly expressing the image semantics.  This feature
vector is taken as input by a hierarchical recurrent neural network composed of
two levels: a sentence RNN and a word RNN.  The sentence RNN receives the image
features, decides how many sentences to generate in the resulting paragraph,
and produces an input topic vector for each sentence.  Given this topic
vector, the word RNN generates the words of a single sentence.  We also show
how to transfer knowledge from a dense image
captioning~\cite{johnson2016densecap} task to our model for paragraph
generation.

\subsection{Region Detector}
The region detector receives an input image of size $3\times H\times W$,
detects regions of interest, and produces a feature vector of dimension
$D=4096$ for each region. Our region
detector follows \cite{ren2015faster,johnson2016densecap}; 
we provide a summary here for completeness:
The image is resized so that its longest edge is 720 pixels, and is then
passed through a convolutional network initialized from 
the 16-layer VGG network~\cite{simonyan2015very}. The resulting
feature map is processed by a region proposal network~\cite{ren2015faster},
which regresses from a set of anchors to propose regions of
interest. These regions are projected onto the convolutional feature 
map, and the corresponding region of the feature map is reshaped
to a fixed size using bilinear interpolation
and processed by two fully-connected layers to give a vector of dimension $D$
for each region.

Given a dataset of images and ground-truth regions of interest, the region
detector can be trained in an end-to-end fashion as in \cite{ren2015faster} for
object detection and \cite{johnson2016densecap} for dense captioning. Since
paragraph descriptions do not have annotated groundings to regions of interest,
we use a region detector trained for dense image captioning on the Visual
Genome dataset~\cite{krishnavisualgenome}, using the publicly available
implementation of \cite{johnson2016densecap}. This produces $M=50$ detected
regions.

One alternative worth noting is to use a region detector trained strictly for
object detection, rather than dense captioning. Although such an approach would
capture many salient objects in an image, its paragraphs would suffer:
an ideal paragraph describes not only objects, but also scenery and
relationships, which are better captured by dense captioning task that captures
\emph{all} noteworthy elements of a scene.

\subsection{Region Pooling}
The region detector produces a set of vectors
$v_1,\ldots,v_M\in\mathbb{R}^{D}$, each describing a different region in the input image.
We wish to aggregate these vectors into a single pooled vector
$v_p\in\mathbb{R}^P$ that compactly describes the content of the image.
To this end, we learn a projection matrix $W_{pool}\in\mathbb{R}^{P\times D}$
and bias $b_{pool}\in\mathbb{R}^P$; the pooled vector $v_p$ is computed by
projecting each region vector using $W_{pool}$ and taking an elementwise maximum,
so that $v_p=\max_{i=1}^M(W_{pool}v_i + b_{pool})$.
While alternative approaches for representing collections of regions, such as
spatial attention~\cite{xu2015show}, may also be possible, we view these as
complementary to the model proposed in this paper; furthermore we note recent
work~\cite{qi2016pointnet} which has proven max pooling sufficient for
representing any continuous set function, giving motivation that max pooling
does not, in principle, sacrifice expressive power.

\subsection{Hierarchical Recurrent Network}
The pooled region vector $v_p\in\mathbb{R}^P$ is given as input to a hierarchical
neural language model composed of two modules: a \emph{sentence RNN}
and a \emph{word RNN}. The sentence RNN is responsible for deciding the number
of sentences $S$ that should be in the generated paragraph and for producing
a $P$-dimensional \emph{topic vector} for each of these sentences. Given a topic
vector for a sentence, the word RNN generates the words of that sentence.
We adopt the standard LSTM architecture~\cite{hochreiter1997long} for both the
word RNN and sentence RNN.

As an alternative to this hierarchical approach, one could instead use a
non-hierarchical language model to directly generate the words of a
paragraph, treating the end-of-sentence token as another word in the
vocabulary. Our hierarchical model is advantageous because it reduces the
length of time over which the recurrent networks must reason. Our paragraphs
contain an average of 67.5 words (Tab.~\ref{tab:data_stats}), so a
non-hierarchical approach must reason over dozens of time steps, which is
extremely difficult for language models.  However, since our paragraphs contain
an average of 5.7 sentences, each with an average of 11.9 words, both the
paragraph and sentence RNNs need only reason over much shorter time-scales,
making learning an appropriate representation much more tractable.

\vspace{-3mm}
\paragraph{Sentence RNN}
The sentence RNN is a single-layer LSTM with hidden size $H=512$ and initial
hidden and cell states set to zero.
At each time step, the sentence RNN receives the pooled region vector $v_p$ as input,
and in turn produces a sequence of hidden states $h_1,\ldots,h_S\in\mathbb{R}^H$,
one for each sentence in the paragraph.
Each hidden state $h_i$ is used in two ways: First, a linear
projection from $h_i$ and a logistic classifier produce a
distribution $p_i$ over the two states $\{\texttt{CONTINUE}=0, \texttt{STOP}=1\}$
which determine whether the $i$th sentence is the last sentence in the paragraph.
Second, the hidden state $h_i$ is fed through a two-layer fully-connected
network to produce the topic vector
$t_i\in\mathbb{R}^P$ for the $i$th sentence of the paragraph, which is the
input to the word RNN.

\paragraph{Word RNN} The word RNN is a two-layer LSTM with hidden size $H=512$,
which, given a topic vector $t_i\in\mathbb{R}^{P}$ from the sentence RNN, is
responsible for generating the words of a sentence.  We follow the input
formulation of~\cite{vinyals2015show}: the first and second inputs to the RNN
are the topic vector and a special \texttt{START} token, and subsequent inputs
are learned embedding vectors for the words of the sentence. At each timestep
the hidden state of the last LSTM layer is used to predict a distribution over
the words in the vocabulary, and a special \texttt{END} token signals the end
of a sentence.  After each Word RNN has generated the words of their respective
sentences, these sentences are finally concatenated to form the generated
paragraph.

\subsection{Training and Sampling}
Training data consists of pairs $(x, y)$, with $x$ an image and $y$ a
ground-truth paragraph description for that image, where $y$ has $S$ sentences,
the $i$th sentence has $N_i$ words, and $y_{ij}$ is the $j$th word of the
$i$th sentence. After computing the pooled region vector $v_p$ for the image,
we unroll the sentence RNN for $S$ timesteps, giving a distribution $p_i$ over
the $\{\texttt{CONTINUE}, \texttt{STOP}\}$ states for each sentence. We feed
the sentence topic vectors to $S$ copies of the word RNN, unrolling the $i$th
copy for $N_i$ timesteps, producing distributions $p_{ij}$ over each word of each
sentence. Our training loss $\ell(x, y)$ for the example $(x, y)$ is a
weighted sum of two cross-entropy terms: a \emph{sentence loss}
$\ell_{sent}$ on the stopping distribution $p_i$,
and a \emph{word loss} $\ell_{word}$ on the word distribution $p_{ij}$:
\vspace{-5mm}

\begin{align}
  \ell(x, y) =
  &\lambda_{sent}\sum_{i=1}^S
      \ell_{sent}(p_i, \mathbf{I}\left[i = S\right])\\
      + &\lambda_{word}\sum_{i=1}^S\sum_{j=1}^{N_i}
      \ell_{word}(p_{ij}, y_{ij})
\end{align}

To generate a paragraph for an image, we run the sentence RNN forward until the
stopping probability $p_i(\texttt{STOP})$ exceeds a threshold $T_{\texttt{STOP}}$
or after $S_{MAX}$ sentences, whichever comes first. We then sample sentences
from the word RNN, choosing the most likely word at each timestep and stopping
after choosing the \texttt{STOP} token or after $N_{MAX}$ words. We set the
parameters \mbox{$T_{\texttt{STOP}}=0.5$}, \mbox{$S_{MAX}=6$}, and \mbox{$N_{MAX}=50$} based on
validation set performance.

\subsection{Transfer Learning}
\vspace{-1mm}
Transfer learning has become pervasive in computer vision.  For tasks such as
object detection~\cite{ren2015faster} and image
captioning~\cite{donahue2015long,karpathy2015deep,vinyals2015show,xu2015show},
it has become standard practice not only to process images with convolutional
neural networks, but also to initialize the weights of these networks from
weights that had been tuned for image classification, such as the 16-layer VGG
network~\cite{simonyan2015very}.  Initializing from a pre-trained convolutional
network allows a form of knowledge transfer from large classification datasets,
and is particularly effective on datasets of limited size.
Might transfer learning also be useful for paragraph generation?

We propose to utilize transfer learning in two ways. First, we initialize our
region detection network from a model trained for dense image
captioning~\cite{johnson2016densecap}; although our model is end-to-end
differentiable, we keep this sub-network fixed during training both for
efficiency and also to prevent overfitting.  Second, we initialize the
word embedding vectors, recurrent network weights, and output linear projection
of the word RNN from a language model that had been trained on region-level
captions~\cite{johnson2016densecap}, fine-tuning these parameters during
training to be better suited for the task of paragraph generation.  Parameters
for tokens not present in the region model are initialized from the parameters
for the \texttt{UNK} token.  This initialization strategy allows our model to
utilize linguistic knowledge learned on large-scale region caption
datasets~\cite{krishnavisualgenome} to produce better paragraph descriptions,
and we validate the efficacy of this strategy in our experiments.

%% file: experiments.tex
In this section we describe our paragraph generation experiments on
the collected data described
in Sec.~\ref{sec:paragraphs}, which we divide into
14,575 training, 2,487 validation, and 2,489 testing images.

\begin{table*}[t]
\begin{tabular}{lcccccc}
  \hline
  & METEOR & CIDEr & BLEU-1 & BLEU-2 & BLEU-3 & BLEU-4 \\
  \hline
  Sentence-Concat & 12.05 & 6.82 &  31.11 & 15.10 & 7.56 & 3.98\\
  Template & 14.31 & 12.15 &  37.47 & 21.02 & 12.30 & 7.38\\
  DenseCap-Concat & 12.66 & 12.51 &  33.18 & 16.92 & 8.54 & 4.54\\
  Image-Flat (\cite{karpathy2015deep}) & 12.82 & 11.06 &  34.04 & 19.95 & 12.20 & 7.71 \\
  Regions-Flat-Scratch & 13.54 & 11.14 & 37.30 & 21.70 & 13.07 & 8.07 \\
  Regions-Flat-Pretrained & 14.23 & 12.13 & 38.32 & 22.90 & 14.17 & \textbf{8.97} \\
  Regions-Hierarchical (ours)\hspace{-3mm} & \textbf{15.95} & \textbf{13.52} & \textbf{41.90} & \textbf{24.11} & \textbf{14.23} & 8.69 \\
  \hline
  Human & 19.22 & 28.55 &  42.88 & 25.68 & 15.55 & 9.66\\
  \hline
\end{tabular}
\caption{Main results for generating paragraphs.
Our Region-Hierarchical method is compared with six baseline models and human performance along six language metrics.
\vspace{-2mm}}
  \label{tab:main_results}
\end{table*}

\subsection{Baselines}
\paragraph{Sentence-Concat:} To demonstrate the difference between
sentence-level and paragraph captions, this baseline samples and concatenates
five sentence captions from a model~\cite{karpathy2015deep} trained on MS COCO
captions~\cite{lin2014microsoft}. The first sentence uses beam search (beam
size $=2$) and the rest are sampled. The motivation for this is as follows: the
image captioning model first produces the sentence that best describes the
image as a whole, and subsequent sentences use
sampling in order to generate a diverse range of sentences, since the
alternative is to repeat the same sentence from beam search.  We have validated
that this approach works better than using either only beam search or only
sampling, as the intent is to make the strongest possible comparison at a
task-level to standard image captioning.  We also note that, while
Sentence-Concat is trained on MS COCO, all images in our dataset are also in MS
COCO, and our descriptions were also written by users on Amazon Mechanical Turk.

\paragraph{Image-Flat:} This model uses a flat representation for both images
and language, and is equivalent to the standard image captioning model
NeuralTalk~\cite{karpathy2015deep}. It takes the whole image as input,
and decodes into a paragraph token by token.  We use the publically available
implementation of \cite{karpathy2015deep}, which uses the 16-layer VGG
network~\cite{simonyan2015very} to extract CNN features and projects them as
input into an LSTM~\cite{hochreiter1997long}, training the whole model jointly
end-to-end.

\paragraph{Template:}
This method represents a very different approach to generating paragraphs,
similar in style to an open-world version of more classical methods like
BabyTalk~\cite{kulkarni2011baby}, which converts a structured representation of
an image into text via a handful of manually specified templates.  The first
step of our template-based baseline is to detect and describe many regions in a
given target image using a pre-trained dense captioning
model~\cite{johnson2016densecap}, which produces a set of region descriptions
tied with bounding boxes and detection scores.  The region descriptions are
parsed into a set of subjects, verbs, objects, and various modifiers according
to part of speech tagging and a handful of
TokensRegex~\cite{chang2014tokensregex} rules, which we find suffice to parse
the vast majority ($\geq 99\%$) of the fairly simplistic and short region-level descriptions.

Each parsed word is scored by the sum of its detection score and the log probability of the generated tokens in the original region description.
Words are then merged into a coherent graph representing the scene, where each node combines all words with the same text and overlapping bounding boxes.
Finally, text is generated using the top $N = 25$ scored nodes, prioritizing \verb|subject-verb-object| triples first in generation, and representing all other nodes with existential ``there is/are'' statements.

\paragraph{DenseCap-Concat:}
This baseline is similar to Sentence-Concat, but instead concatenates
DenseCap~\cite{johnson2016densecap} predictions as separate sentences in order
to form a paragraph. The intent of analyzing this method is to disentangle two
key parts of the Template method: captioning and detection (\ie DenseCap), and
heuristic recombination into paragraphs.
We combine the top $n=14$ outputs of DenseCap to
form DenseCap-Concat's output based on validation CIDEr+METEOR.

\paragraph{Other Baselines:}
``Regions-Flat-Scratch'' uses a flat language model for decoding and initializes it from scratch.
The language model input is the projected and pooled region-level image features.
``Regions-Flat-Pretrained'' uses a pre-trained language model.
These baselines are included to show the benefits of decomposing the image into regions and pre-training the language model.

\subsection{Implementation Details}
All baseline neural language models use two layers of
LSTM~\cite{hochreiter1997long} units with 512 dimensions.  The feature pooling
dimension $P$ is 1024, and we set $\lambda_{sent}=5.0$ and $\lambda_{word}=1.0$
based on validation set performance.  Training is done via stochastic gradient
descent with Adam~\cite{kingma2015adam}, implemented in
Torch.  Of critical note is that model checkpoint selection is based on the
best combined METEOR and CIDEr score on the validation set -- although models
tend to minimize validation loss fairly quickly, it takes much longer training
for METEOR and CIDEr scores to stop improving.

\subsection{Main Results}

\begin{figure*}
  \includegraphics[width=0.90\textwidth]{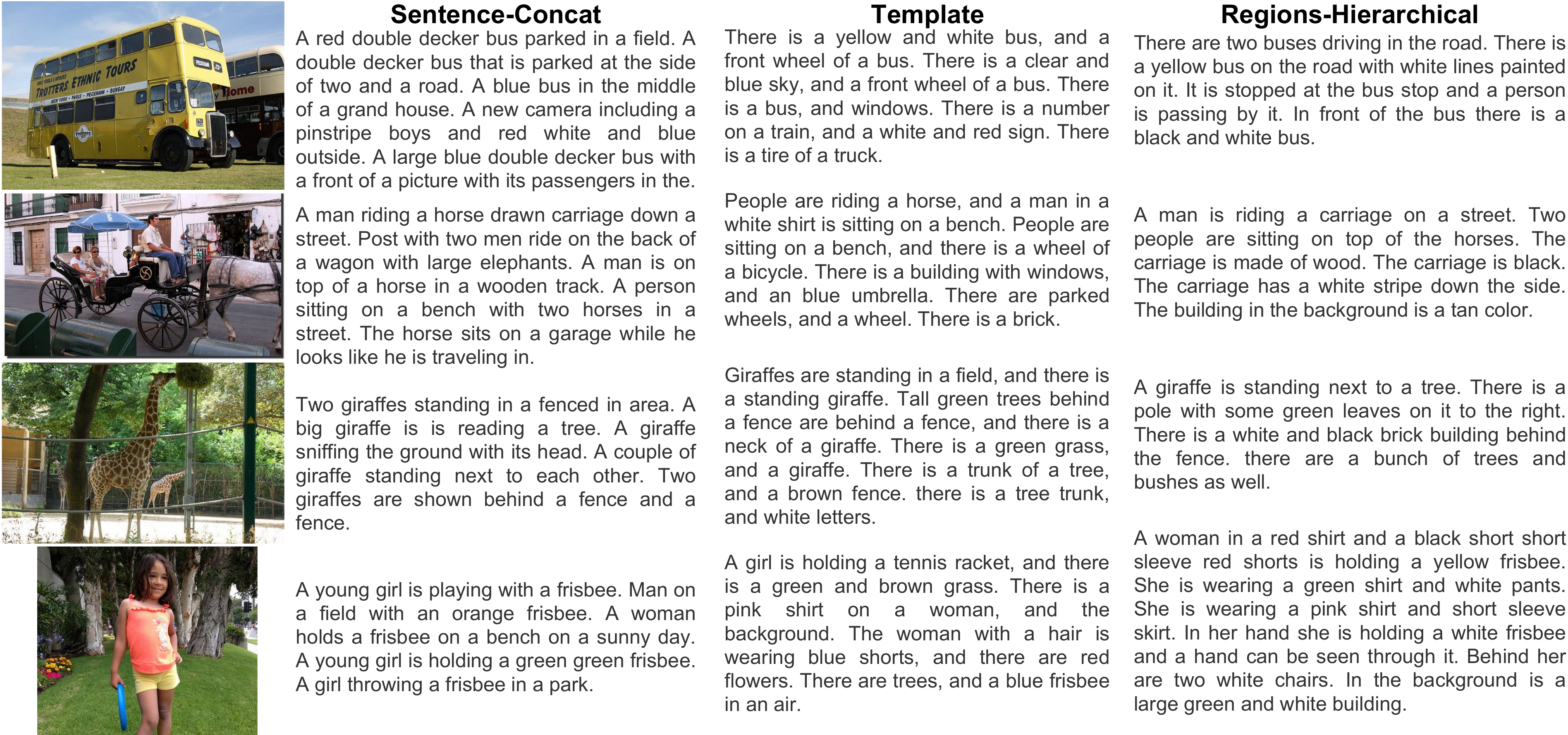}
  \caption{
    Example paragraph generation results for our model
  (Regions-Hierarchical) and the Sentence-Concat and Template
baselines.  The first three rows are positive results and the last row is a
failure case.}
  \label{fig:qualitative}
\end{figure*}

We present our main results at generating paragraphs in Tab.~\ref{tab:main_results}, which are evaluated across six language metrics: CIDEr~\cite{vedantam2015cider}, METEOR~\cite{denkowski2014meteor}, and BLEU-\{1,2,3,4\}~\cite{papineni2002bleu}.
The Sentence-Concat method performs poorly, achieving the lowest scores across all metrics. Its lackluster performance provides further evidence of the stark differences between single-sentence captioning and paragraph generation.
Surprisingly, the hard-coded template-based approach performs reasonably well, particularly on CIDEr, METEOR, and BLEU-1, where it is competitive with some of the neural approaches.
This makes sense: the template approach is provided with a strong prior about image content since it receives region-level captions~\cite{johnson2016densecap} as input, and the many expletive ``there is/are'' statements it makes, though uninteresting, are safe, resulting in decent scores.
However, its relatively poor performance on BLEU-3 and BLEU-4 highlights the limitation of reasoning about regions in isolation -- it is unable to produce much text relating regions to one another, and further suffers from a lack of ``connective tissue'' that transforms paragraphs from a series of disconnected thoughts into a coherent whole.
DenseCap-Concat scores worse than Template on all metrics except CIDEr, illustrating the necessity of Template's caption parsing and recombination.

Image-Flat, trained on the task of paragraph generation, outperforms
Sentence-Concat, and the region-based reasoning of Regions-Flat-Scratch
improves results further on all metrics.  Pre-training results in improvements
on all metrics, and our full model, Regions-Hierarchical, achieves the highest
scores among all methods on every metric except BLEU-4.  One hypothesis for the
mild superiority of Regions-Flat-Pretrained on BLEU-4 is that it is better able
to reproduce words immediately at the end and beginning of sentences more
exactly due to their non-hierarchical structure, providing a slight boost in
BLEU scores.

To make these metrics more interpretable, we performed a human evaluation by
collecting an additional paragraph for 500 randomly chosen images, with results
in the last row of Tab.~\ref{tab:main_results}.  As expected, humans produce
superior descriptions to any automatic method, performing better on all
language metrics considered.  Of particular note is the large gap between
humans our the best model on CIDEr and METEOR, which are both designed to
correlate well with human
judgment~\cite{vedantam2015cider,denkowski2014meteor}.

Finally, we note that we have also tried the SPICE
evaluation metric~\cite{anderson2016spice}, which has shown to correlate well with human
judgements for sentence-level image captioning. Unfortunately, SPICE does not seem well-suited
for evaluating long paragraph descriptions -- it does not handle coreference
or distinguish between different instances of the same object category. These
are reasonable design decisions for sentence-level captioning, but is
less applicable to paragraphs.
%SPICE also does not measure grammatical or syntactic correctness, which is captured to some extent by metrics based on n-grams.
In fact, human paragraphs achieved a considerably lower SPICE score than
automated methods.

\subsection{Qualitative Results}
\vspace{-0mm}

We present qualitative results from our model and the Sentence-Concat and Template baselines in Fig.~\ref{fig:qualitative}.
Some interesting properties of our model's predictions include its use of
coreference in the first example (``a bus'', ``it'', ``the bus'') and its
ability to capture relationships between objects in the second example.
Also of note is the order in which our model chooses to describe the image:
the first sentence tends to be fairly high level, middle sentences give some
details about scene elements mentioned earlier in the description, and the
last sentence often describes something in the background, which other methods
are not able to capture. Anecdotally, we observed that this follows the same
order with which most humans tended to describe images.

The failure case in the last row highlights another interesting phenomenon:
even though our model was wrong about the semantics of the image,
calling the girl ``a woman'', it has learned that ``woman'' is consistently associated with
female pronouns (``she'', ``she'', ``her hand'', ``behind her'').

It is also worth noting the general behavior of the two baselines.  Paragraphs
from Sentence-Concat tend to be repetitive in sentence structure and are often
simply inaccurate due to the sampling required to generate multiple sentences.
On the other hand, the Template baseline is largely accurate, but has
uninteresting language and lacks the ability to determine which things are most
important to describe.  In contrast, Regions-Hierarchical stays relevant
and furthermore exhibits more interesting patterns of language.

\begin{table*}[t]
\begin{tabular}{lcccccccc}
  %& \multiline{Sentence\\-Concat} & Template & \multiline{Image\\-Flat} & \multiline{Regions\\-Flat\\-Scratch} & \multiline{Regions\\-Flat\\-Pretrained} & \multiline{Regions\\-Hierarchical} & Human \\
  \hline
& \multiline{Average\\Length} & \multiline{Std. Dev.\\Length} & Diversity & Nouns & Verbs & Pronouns & \multiline{Vocab\\Size} \\
  \hline
  Sentence-Concat & 56.18 & 4.74 & 34.23 & 32.53 & 9.74 & 0.95 & 2993 \\
  Template & 60.81 & 7.01 & 45.42 & 23.23 & 11.83 & 0.00 & 422 \\
  Regions-Hierarchical\hspace{-2mm} & 70.47 & 17.67 & 40.95 & 24.77 & 13.53 & 2.13 & 1989 \\
  Human & 67.51 & 25.95 & 69.92 & 25.91 & 14.57 & 2.42 & 4137 \\
  \hline
\end{tabular}
\caption{Language statistics of test set predictions.
Part of speech statistics are given as percentages, and diversity is calculated as in Section~\ref{sec:paragraphs}.
``Vocab Size'' indicates the number of unique tokens output across the entire test set, and human numbers are calculated from ground truth.
Note that the diversity score for humans differs slightly from the score in Tab.~\ref{tab:data_stats}, which is calculated on the entire dataset.
}
  \label{tab:output_language_stats}
\end{table*}

\begin{figure*}
  \centering
  \includegraphics[width=0.85\textwidth]{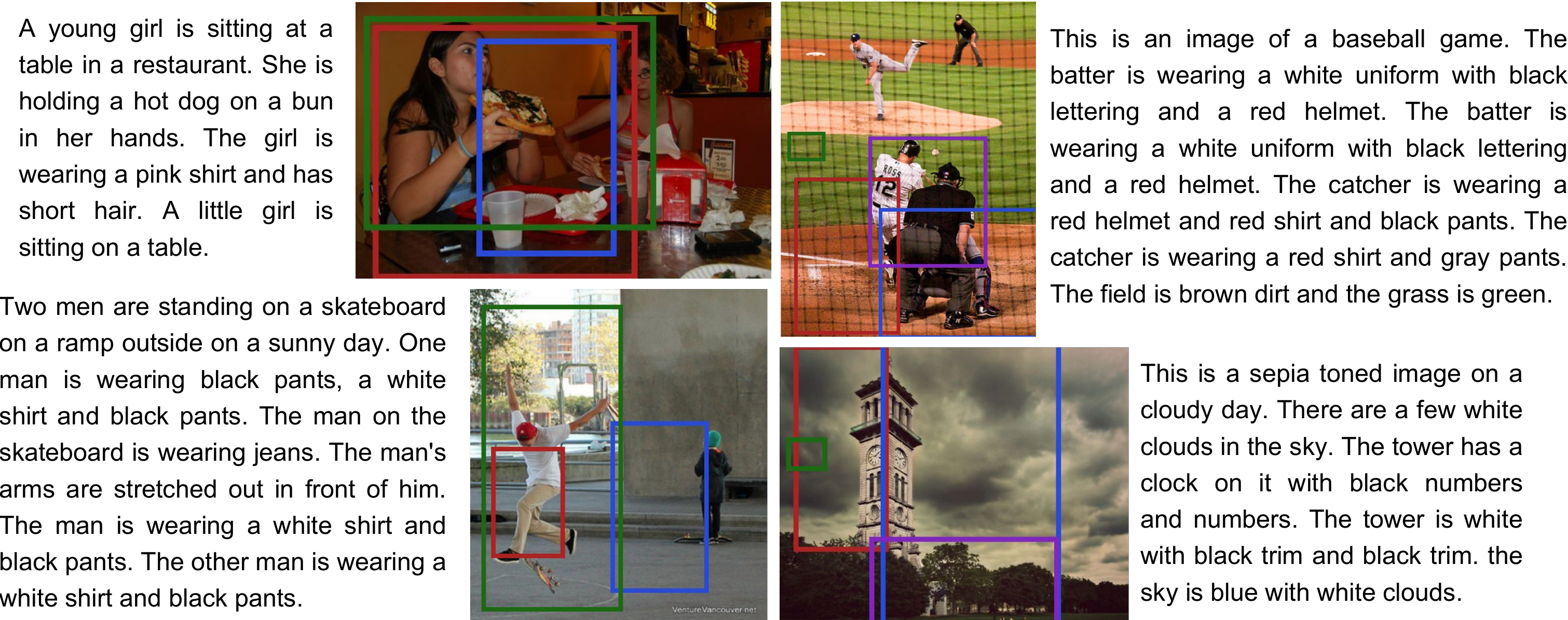}
  \caption{Examples of paragraph generation from only a few regions.
		Since only a small number of regions are used, this data is extremely out
		of sample for the model, but it is still able to focus on the regions of
		interest while ignoring the rest of the image.
		\vspace{-4mm}}
  \label{fig:regions}
\end{figure*}

\subsection{Paragraph Language Analysis}
To shed a quantitative light on the linguistic phenomena generated, in
Tab.~\ref{tab:output_language_stats} we show statistics of the language
produced by a representative spread of methods.

Our hierarchical approach generates text of similar average length and variance
as human descriptions, with Sentence-Concat and the Template approach somewhat
shorter and less varied in length.  Sentence-Concat is also the least diverse
method, though all automatic methods remain far less diverse than human
sentences, indicating ample opportunity for improvement. According to this
diversity metric, the Template approach is actually the
most diverse automatic method, which may be attributed to how the method is
hard-coded to sequentially describe each region in the scene in turn,
regardless of importance or how interesting such an output may be (see
Fig.~\ref{fig:qualitative}).  While both our hierarchical approach and the
Template method produce text with similar portions of nouns and verbs as human
paragraphs, only our approach was able to generate a reasonable quantity of
pronouns.  Our hierarchical method also had a much wider vocabulary compared to
the Template approach, though Sentence-Concat, trained on hundreds of thousands
of MS COCO~\cite{lin2014microsoft} captions, is a bit larger.

\subsection{Generating Paragraphs from Fewer Regions}

As an exploratory experiment in order to highlight the interpretability of our
model, we investigate generating paragraphs from a smaller number of regions
than the $M=50$ used in the rest of this work.  Instead, we only give our
method access to the top few detected regions as input, with the hope that the
generated paragraph focuses only on those particularly regions, preferring not
to describe other parts of the image.  The results for a handful of images are
shown in Fig.~\ref{fig:regions}.  Although the input is extremely out of sample
compared to the training data, the results are still quite reasonable -- the
model generates paragraphs describing the detected regions without much mention
of objects or scenery outside of the detections.  Taking the top-right image as
an example, despite a few linguistic mistakes, the paragraph generated by our
model mentions the batter, catcher, dirt, and grass, which all appear in the
top detected regions, but does not pay heed to the pitcher or the umpire in the
background.

%% file: discussion.tex
\vspace{-2mm}
In this paper we have introduced the task of describing images with long,
descriptive paragraphs, and presented a hierarchical approach for
generation that leverages the compositional structure of both images and
language.  We have shown that paragraph generation is different from
traditional image captioning and have tailored our model to suit these
differences.  Experimentally, we have demonstrated the advantages of our
approach over traditional image captioning methods and shown how region-level
knowledge can be effectively transferred to paragraph captioning.  We
have also demonstrated the benefits of our model in interpretability,
generating descriptive paragraphs using only a subset of image regions.
We anticipate further opportunities for knowledge transfer at the
intersection of vision and language, and project that visual and lingual
compositionality will continue to lie at the heart of effective paragraph
generation.

%% file: paper.bbl
\begin{thebibliography}{10}\itemsep=-1pt

\bibitem{anderson2016spice}
P.~Anderson, B.~Fernando, M.~Johnson, and S.~Gould.
\newblock Spice: Semantic propositional image caption evaluation.
\newblock In {\em European Conference on Computer Vision}, pages 382--398.
  Springer, 2016.

\bibitem{chan2016listen}
W.~Chan, N.~Jaitly, Q.~V. Le, and O.~Vinyals.
\newblock Listen, attend, and spell: A neural network for large vocabulary
  conversational speech recognition.
\newblock In {\em ICASSP}, 2016.

\bibitem{chang2014tokensregex}
A.~X. Chang and C.~D. Manning.
\newblock Tokensregex: Defining cascaded regular expressions over tokens.
\newblock Technical report, CSTR 2014-02, Department of Computer Science,
  Stanford University, 2014.

\bibitem{chen2015mind}
X.~Chen and C.~Lawrence~Zitnick.
\newblock Mind's eye: A recurrent visual representation for image caption
  generation.
\newblock In {\em CVPR}, 2015.

\bibitem{denkowski2014meteor}
M.~Denkowski and A.~Lavie.
\newblock Meteor universal: Language specific translation evaluation for any
  target language.
\newblock In {\em EACL Workshop on Statistical Machine Translation}, 2014.

\bibitem{donahue2015long}
J.~Donahue, L.~Anne~Hendricks, S.~Guadarrama, M.~Rohrbach, S.~Venugopalan,
  K.~Saenko, and T.~Darrell.
\newblock Long-term recurrent convolutional networks for visual recognition and
  description.
\newblock In {\em CVPR}, 2015.

\bibitem{el1995hierarchical}
S.~El~Hihi and Y.~Bengio.
\newblock Hierarchical recurrent neural networks for long-term dependencies.
\newblock In {\em NIPS}, 1995.

\bibitem{farhadi2010every}
A.~Farhadi, M.~Hejrati, M.~A. Sadeghi, P.~Young, C.~Rashtchian, J.~Hockenmaier,
  and D.~Forsyth.
\newblock Every picture tells a story: Generating sentences from images.
\newblock In {\em ECCV}, 2010.

\bibitem{hochreiter1997long}
S.~Hochreiter and J.~Schmidhuber.
\newblock Long short-term memory.
\newblock {\em Neural computation}, 1997.

\bibitem{hodosh2013framing}
M.~Hodosh, P.~Young, and J.~Hockenmaier.
\newblock Framing image description as a ranking task: Data, models and
  evaluation metrics.
\newblock {\em Journal of Artificial Intelligence Research}, 47:853--899, 2013.

\bibitem{johnson2016densecap}
J.~Johnson, A.~Karpathy, and L.~Fei-Fei.
\newblock {DenseCap}: Fully convolutional localization networks for dense
  captioning.
\newblock In {\em CVPR}, 2016.

\bibitem{karpathy2015deep}
A.~Karpathy and L.~Fei-Fei.
\newblock Deep visual-semantic alignments for generating image descriptions.
\newblock In {\em CVPR}, 2015.

\bibitem{karpathy2014deep}
A.~Karpathy, A.~Joulin, and L.~Fei-Fei.
\newblock Deep fragment embeddings for bidirectional image sentence mapping.
\newblock In {\em NIPS}, 2014.

\bibitem{kingma2015adam}
D.~Kingma and J.~Ba.
\newblock Adam: A method for stochastic optimization.
\newblock In {\em ICLR}, 2015.

\bibitem{koutnik2014clockwork}
J.~Koutnik, K.~Greff, F.~Gomez, and J.~Schmidhuber.
\newblock A clockwork {RNN}.
\newblock In {\em ICML}, 2014.

\bibitem{krishnavisualgenome}
R.~Krishna, Y.~Zhu, O.~Groth, J.~Johnson, K.~Hata, J.~Kravitz, S.~Chen,
  Y.~Kalantidis, L.-J. Li, D.~A. Shamma, M.~Bernstein, and L.~Fei-Fei.
\newblock Visual genome: Connecting language and vision using crowdsourced
  dense image annotations.
\newblock {\em arXiv preprint arXiv:1602.07332}, 2016.

\bibitem{kulkarni2011baby}
G.~Kulkarni, V.~Premraj, S.~Dhar, S.~Li, Y.~Choi, A.~C. Berg, and T.~L. Berg.
\newblock Baby talk: Understanding and generating image descriptions.
\newblock In {\em CVPR}, 2011.

\bibitem{li2015hierarchical}
J.~Li, M.-T. Luong, and D.~Jurafsky.
\newblock A hierarchical neural autoencoder for paragraphs and documents.
\newblock In {\em ACL}, 2015.

\bibitem{lin2015hierarchical}
R.~Lin, S.~Liu, M.~Yang, M.~Li, M.~Zhou, and S.~Li.
\newblock Hierarchical recurrent neural network for document modeling.
\newblock In {\em EMNLP}, 2015.

\bibitem{lin2014microsoft}
T.-Y. Lin, M.~Maire, S.~Belongie, J.~Hays, P.~Perona, D.~Ramanan,
  P.~Doll{\'a}r, and C.~L. Zitnick.
\newblock Microsoft coco: Common objects in context.
\newblock In {\em ECCV}, 2014.

\bibitem{manning2014stanford}
C.~D. Manning, M.~Surdeanu, J.~Bauer, J.~R. Finkel, S.~Bethard, and
  D.~McClosky.
\newblock The stanford {CoreNLP} natural language processing toolkit.
\newblock In {\em ACL (System Demonstrations)}, pages 55--60, 2014.

\bibitem{mao2014deep}
J.~Mao, W.~Xu, Y.~Yang, J.~Wang, Z.~Huang, and A.~Yuille.
\newblock Deep captioning with multimodal recurrent neural networks (m-{RNN}).
\newblock {\em ICLR}, 2015.

\bibitem{marcus1993building}
M.~P. Marcus, M.~A. Marcinkiewicz, and B.~Santorini.
\newblock Building a large annotated corpus of english: The penn treebank.
\newblock {\em Computational linguistics}, 19(2):313--330, 1993.

\bibitem{papineni2002bleu}
K.~Papineni, S.~Roukos, T.~Ward, and W.-J. Zhu.
\newblock Bleu: a method for automatic evaluation of machine translation.
\newblock In {\em ACL}, pages 311--318. Association for Computational
  Linguistics, 2002.

\bibitem{qi2016pointnet}
C.~R. Qi, H.~Su, K.~Mo, and L.~J. Guibas.
\newblock Pointnet: Deep learning on point sets for 3d classification and
  segmentation.
\newblock {\em arXiv preprint arXiv:1612.00593}, 2016.

\bibitem{ren2015faster}
S.~Ren, K.~He, R.~Girshick, and J.~Sun.
\newblock Faster r-cnn: Towards real-time object detection with region proposal
  networks.
\newblock In {\em NIPS}, 2015.

\bibitem{rohrbach2014coherent}
A.~Rohrbach, M.~Rohrbach, W.~Qiu, A.~Friedrich, M.~Pinkal, and B.~Schiele.
\newblock Coherent multi-sentence video description with variable level of
  detail.
\newblock In {\em German Conference on Pattern Recognition}, pages 184--195.
  Springer, 2014.

\bibitem{simonyan2015very}
K.~Simonyan and A.~Zisserman.
\newblock Very deep convolutional networks for large-scale image recognition.
\newblock In {\em ICLR}, 2015.

\bibitem{vedantam2015cider}
R.~Vedantam, C.~Lawrence~Zitnick, and D.~Parikh.
\newblock {CIDEr}: Consensus-based image description evaluation.
\newblock In {\em CVPR}, 2015.

\bibitem{vinyals2015show}
O.~Vinyals, A.~Toshev, S.~Bengio, and D.~Erhan.
\newblock Show and tell: A neural image caption generator.
\newblock In {\em CVPR}, 2015.

\bibitem{xu2015show}
K.~Xu, J.~Ba, R.~Kiros, K.~Cho, A.~Courville, R.~Salakhudinov, and Y.~Bengio.
\newblock Show, attend, and tell: Neural image caption generation with visual
  attention.
\newblock In {\em ICML}, 2015.

\bibitem{yao2015describing}
L.~Yao, A.~Torabi, K.~Cho, N.~Ballas, C.~Pal, H.~Larochelle, and A.~Courville.
\newblock Describing videos by exploiting temporal structure.
\newblock In {\em ICCV}, 2015.

\bibitem{you2016image}
Q.~You, H.~Jin, Z.~Wang, C.~Fang, and J.~Luo.
\newblock Image captioning with semantic attention.
\newblock In {\em CVPR}, 2016.

\bibitem{young2014image}
P.~Young, A.~Lai, M.~Hodosh, and J.~Hockenmaier.
\newblock From image descriptions to visual denotations: New similarity metrics
  for semantic inference over event descriptions.
\newblock {\em Transactions of the Association for Computational Linguistics},
  2:67--78, 2014.

\bibitem{yu2016video}
H.~Yu, J.~Wang, Z.~Huang, Y.~Yang, and W.~Xu.
\newblock Video paragraph captioning using hierarchical recurrent neural
  networks.
\newblock In {\em CVPR}, 2016.

\end{thebibliography}
